\documentclass[conference]{IEEEtran}

\makeatletter

\def\ps@IEEEtitlepagestyle{%
  \def\@oddfoot{\mycopyrightnotice}%
  \def\@evenfoot{}%
}
\def\mycopyrightnotice{%
  {\footnotesize XXX-X-XXXX-XXXX-X/XX/\$XX.00~\copyright~20XX IEEE\hfill}
  \gdef\mycopyrightnotice{}
}

\usepackage{blindtext}
\usepackage{eso-pic}
\IEEEoverridecommandlockouts
\usepackage{cite}
\usepackage{amsmath,amssymb,amsfonts}
\usepackage{algorithmic}
\usepackage{graphicx}
\usepackage{textcomp}
\usepackage{xcolor}
\def\BibTeX{{\rm B\kern-.05em{\sc i\kern-.025em b}\kern-.08em
    T\kern-.1667em\lower.7ex\hbox{E}\kern-.125emX}}
    
\usepackage{eso-pic}
\newcommand\AtPageUpperMyright[1]{\AtPageUpperLeft{%
 \put(\LenToUnit{0.17\paperwidth},\LenToUnit{-2cm}){%
     \parbox{0.9\textwidth}{\raggedleft\fontsize{8}{11}\selectfont #1}}%
 }}%
\newcommand{\conf}[1]{%
\AddToShipoutPictureBG*{%
\AtPageUpperMyright{#1}
}
}    

\usepackage{graphicx}
\usepackage[labelfont=bf]{caption}
\usepackage{url}
\usepackage{float}
\usepackage{hyperref}
\usepackage{amsmath}
 
\usepackage{mathtools}
\usepackage{bm, csquotes}
\usepackage{subcaption}
\usepackage{chngcntr}
\counterwithin{figure}{section}
\usepackage{multirow}
\usepackage{array}
\usepackage{enumerate}
\usepackage{tikz}
\usetikzlibrary{arrows.meta, positioning, shapes.geometric}
\usepackage{booktabs}

\usepackage{todonotes}	

\begin{document}
\title{\vspace*{1cm} Unsupervised Electrofacies Classification and Porosity Characterization in the Offshore Keta Basin Using Wireline Logs\\
}

\author{\IEEEauthorblockN{Hamdiya Adams}
\IEEEauthorblockA{Department of Earth Science\\University of Ghana
Legon, Accra, Ghana\\
hamdiya410@gmail.com}
\and 
\IEEEauthorblockN{Theophilus Ansah-Narh}
\IEEEauthorblockA{Ghana Space Science and \\ Technology Institute\\
Accra, Ghana\\
theophilus.ansah-narh@gaec.gov.gh}
\and 
\IEEEauthorblockN{ 
Daniel Kwadwo Asiedu
}
\IEEEauthorblockA{Department of Earth Science\\University of Ghana Legon, 
Accra, Ghana\\
dasiedu@ug.edu.gh}
\and
\IEEEauthorblockN{Bruce Kofi Banoeng-Yakubo}
\IEEEauthorblockA{Department of Earth Science\\University of Ghana Legon,
Accra, Ghana\\
bkbanoeng-yakubo@ug.edu.gh}
\and
\IEEEauthorblockN{Marcellin Atemkeng\textsuperscript{$\dagger$}}
\IEEEauthorblockA{Department of Mathematics\\Rhodes University, Grahamstown, South Africa\\
m.atemkeng@ru.ac.za}
\and
\IEEEauthorblockN{Thomas Armah}
\IEEEauthorblockA{Department of Earth Science\\
University of Ghana\\ Legon, Accra, Ghana\\
tekarmah@gmail.com}
\and
\IEEEauthorblockN{
Richmond Opoku-Sarkodie}
\IEEEauthorblockA{Department of Information Technology and \\ Mathematical Sciences\\Methodist University 
Ghana, Accra, Ghana\\
ropokusarkodie@gmail.com}
\and
\IEEEauthorblockN{Rebecca Davis}
\IEEEauthorblockA{Department of Actuarial Science\\Pentecost University
 Accra, Ghana\\
rdavis@pentvars.edu.gh}
\and
\IEEEauthorblockN{Ezekiel Nii Noye Nortey}
\IEEEauthorblockA{Department of Statistics and \\ Actuarial Science,\\ University of Ghana\\ Legon,
Accra, Ghana\\
ennnortey11@gmail.com}
}


\maketitle
\conf{\textit{ Proc. of the International Conference on Electrical, Computer and Energy Technologies (ICECET) \\ 
06-09 July 2026, Rome-Italy}}

\begin{abstract}
This study presents an unsupervised machine learning workflow for electrofacies analysis in the offshore Keta Basin, Ghana, where core data are scarce. Six standard wireline logs from Well~C were analysed over a depth interval comprising approximately $11{,}195$ samples. K-means clustering was applied in multivariate log space, with the clustering structure evaluated using inertia and silhouette diagnostics. Four clusters were identified, supported by an average silhouette coefficient of approximately $0.50$, indicating moderate but meaningful separation.
The resulting electrofacies exhibit systematic, depth-continuous patterns associated with variations in clay content, porosity, and rock framework properties, forming a geological continuum from shale-dominated to cleaner sandstone-dominated units. The results demonstrate that log-only, unsupervised clustering supported by quantitative metrics provides a robust and reproducible framework for subsurface characterisation. The proposed workflow offers a practical tool for early-stage formation evaluation in frontier offshore basins and a foundation for future integrated studies.
\end{abstract}



\section{Introduction} \label{intro}

Reliable formation evaluation remains a central requirement for subsurface characterisation in frontier offshore basins, where exploration decisions are commonly made under conditions of limited geological control and high operational uncertainty. 
In such settings, wireline logs often constitute the primary source of continuous subsurface information, providing indirect but high-resolution insights into lithology, porosity and fluid-related properties.
This is particularly relevant for emerging offshore provinces, where the acquisition of core data is constrained by drilling costs and logistical challenges, and where early-stage appraisal frequently relies on log-based interpretation workflows.
Within Ghana’s offshore domain, ongoing interest in expanding hydrocarbon resources and improving subsurface knowledge has heightened the need for robust, cost-effective formation evaluation methodologies that can operate effectively in data-scarce environments.

The offshore Keta Basin represents a typical example of such a frontier setting. Despite its strategic location along the Gulf of Guinea and its relevance to Ghana’s broader energy and subsurface development objectives, the basin remains under-characterised relative to more mature neighbouring provinces. Core availability is sparse, and detailed lithological calibration is often unavailable, necessitating a strong dependence on wireline logs for stratigraphic and petrophysical interpretation. These constraints motivate the adoption of quantitative, reproducible approaches that can extract meaningful geological structure from multivariate log responses without requiring labelled training data.

Within this context, electrofacies analysis has emerged as a practical framework for grouping log responses into internally consistent units that reflect underlying lithological and petrophysical similarities.
Electrofacies do not represent lithofacies in the strict sedimentological sense, but rather statistically defined clusters in log-response space that can be interpreted geologically when informed by domain knowledge.
Unsupervised machine learning methods are particularly well-suited to electrofacies analysis in frontier basins, as they do not require pre-existing labels or extensive core control. Early work by \cite{ghosh2016estimation, zhang1999application} demonstrated the potential of machine learning techniques to extract lithologically meaningful patterns from wireline logs, while subsequent studies have shown that unsupervised clustering can provide stable and interpretable electrofacies classifications when geological constraints are applied \cite{hussain2025hybrid, emelyanova2017unsupervised}. 
More recent applications in data-limited and frontier settings further highlight the value of unsupervised approaches for reducing interpreter subjectivity and improving reproducibility, particularly where conventional calibration data are unavailable.

Despite these advances, the application of machine learning-based electrofacies workflows in West African offshore basins remains limited in the published literature. To the authors’ knowledge, no dedicated electrofacies characterisation has previously been reported for the offshore Keta Basin, and few studies have explicitly addressed the integration of unsupervised clustering with quantitative petrophysical estimation in this regional context. This gap is notable given the increasing availability of digital well logs and the growing emphasis on transparent, reproducible analysis pipelines in subsurface studies.

Against this backdrop, the present study develops and applies an unsupervised electrofacies workflow for the offshore Keta Basin based exclusively on standard wireline logs. The contributions of this work are fourfold. First, it provides the first documented electrofacies characterisation of the offshore Keta Basin using unsupervised machine learning. Second, it integrates density–neutron porosity computation with electrofacies analysis to support joint petrophysical and lithological interpretation within a unified framework. Third, the study incorporates geological reasoning alongside quantitative clustering diagnostics to guide the selection and interpretation of electrofacies structures in log-response space. Finally, it presents a fully reproducible, log-only workflow designed for application in data-scarce offshore basins, offering a practical and transparent approach to early-stage formation evaluation where core data are unavailable.

\section{Study Area and Well Data} \label{sec:EDA}
\subsection{Study Area}

The Keta Basin is an offshore sedimentary basin located along the eastern margin of Ghana, extending into the Gulf of Guinea within the West African transform margin system. The basin forms part of a series of pull-apart and marginal basins developed during the opening of the South Atlantic, and is structurally influenced by transform faulting and associated syn-rift to post-rift depositional processes \cite{brownfield2006geology}. 
Compared with Ghana’s more mature offshore provinces, the Keta Basin remains relatively under-explored, with limited publicly available subsurface information and sparse well control. As a result, geological understanding of the basin relies heavily on indirect geophysical and wireline log data.

\begin{figure}
	\begin{minipage}{\linewidth}
		\centering
		\includegraphics[width=1.0\textwidth]{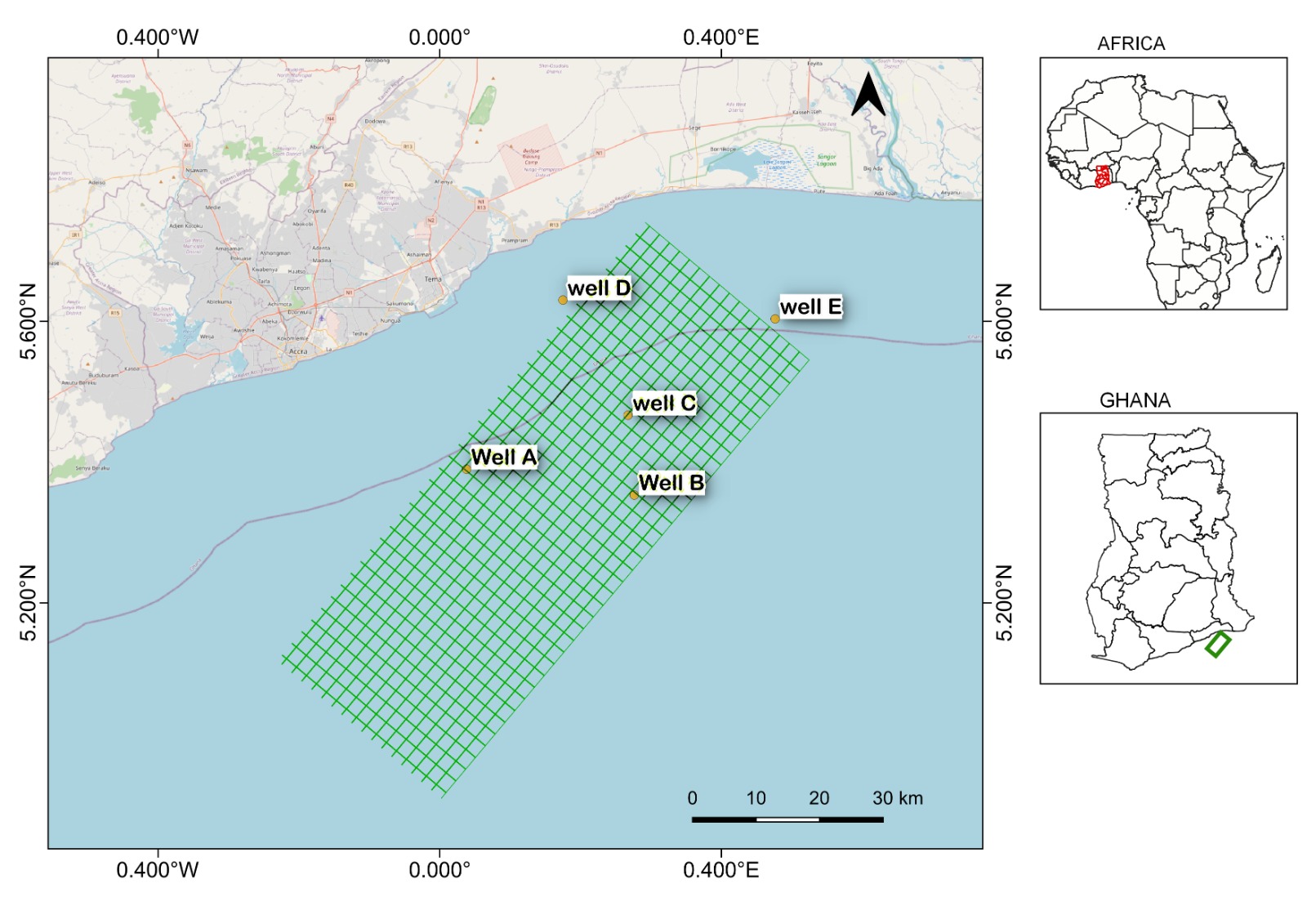} %
	\end{minipage}
	\caption{Study area map showing the offshore Keta Basin along the eastern margin of Ghana. The green gridded polygon outlines the seismic survey coverage, while labelled markers indicate the locations of exploration wells within the study area. The inset maps provide regional context, showing the location of the study area within Ghana and the broader African continent.}
	\label{fig:study_area}
\end{figure}

The present study focuses on Well~C, an offshore exploration well located within the central sector of the Keta Basin (Fig.~\ref{fig:study_area}). The analysed depth interval spans from $1358.34$~m to $3064.31$~m, covering a stratigraphic section characterised by mixed clastic successions typical of marginal marine to shallow marine depositional environments reported for the basin \cite{mascle1987evidence}. This depth range was selected to ensure continuity of the wireline log suite and to exclude shallow sections affected by severe borehole instability. In the absence of core data, Well~C provides a suitable test case for evaluating the applicability of log-based, unsupervised electrofacies analysis in a frontier offshore setting.

\subsection{Data Description}\label{sec:desc}

The dataset analysed in this study consists of approximately 11,195 depth-indexed samples acquired from Well~C in the offshore Keta Basin. The available wireline log suite includes gamma ray (GR), bulk density (RHOZ), neutron porosity (NPHI), compressional sonic transit time (DT), photoelectric factor (PEFZ), and deep resistivity (AHT60). These logs represent a standard open-hole acquisition set routinely used in subsurface formation evaluation and are widely available in frontier offshore settings.

Each log was selected based on its sensitivity to distinct physical properties of the subsurface. Gamma ray records natural radioactivity and is commonly used as a proxy for shale volume in siliciclastic formations. Bulk density and neutron porosity provide independent measurements related to matrix density and hydrogen concentration, respectively, and are frequently analysed together to characterise porosity and lithological effects. The sonic log measures compressional wave transit time and is sensitive to elastic properties and compaction state of the formation. The photoelectric factor responds primarily to mineralogical composition, while deep resistivity provides information related to the .electrical properties of the formation at depth. These measurements form a complementary multivariate dataset suitable for log-based electrofacies analysis \cite{liu2017principles, ellis2007well,rider1996geological}.

Prior to analysis, the wireline data were subjected to quality control and preprocessing to minimise the influence of non-geological artefacts. Depth intervals affected by borehole washout were identified using caliper-derived indicators and removed to ensure reliable log responses. Statistical outliers were screened using a $\pm 3\sigma$ criterion applied independently to each log, a standard approach in wireline log conditioning to suppress spurious measurements while preserving genuine variability \cite{asquith2004basic}. Following quality control, all selected logs were normalised using z-score standardisation to ensure comparable scaling prior to multivariate analysis and to prevent dominance of any single measurement in distance-based clustering.

\section{Methodology}

This section describes the log-based porosity estimation and unsupervised clustering framework adopted for electrofacies characterisation in Well~C of the offshore Keta Basin. The workflow is designed to operate exclusively on standard wireline logs, reflecting the data limitations typical of frontier offshore settings.

\subsection{Porosity Estimation from Wireline Logs}

Porosity was estimated using a combination of density and neutron wireline logs, following conventional petrophysical practice. Density-derived porosity was computed as
\begin{equation}
\phi_D = \frac{\rho_{ma} - \rho_b}{\rho_{ma} - \rho_f},
\end{equation}
where $\rho_{ma}$ is the assumed matrix density, $\rho_b$ is the measured bulk density from the RHOZ log, and $\rho_f$ is the formation fluid density.
A limestone matrix density of $2.71~\text{g\,cm}^{-3}$ and a fluid density of $1.0~\text{g\,cm}^{-3}$ were adopted, consistent with standard assumptions in the absence of core-derived mineralogical calibration \cite{rider1996geological}.

Neutron porosity was obtained directly from the NPHI log,
\begin{equation}
\phi_N = \text{NPHI},
\end{equation}
with the implicit assumption that the log is recorded in limestone units. While this assumption may introduce lithology-dependent bias in mixed clastic intervals, it provides a practical and internally consistent basis for porosity estimation in data-limited settings.

An average porosity was then computed as
\begin{equation}
\phi_{\text{avg}} = \frac{\phi_D + \phi_N}{2},
\end{equation}
which serves as a smoothed porosity indicator that mitigates the individual sensitivities of density and neutron measurements. This averaged porosity was used for descriptive analysis and integration with electrofacies interpretation, rather than as a prediction target.

\subsection{K-Means Clustering Framework}

Electrofacies classification was carried out using the K-means clustering algorithm, which partitions observations in a multivariate feature space by minimising within-cluster variance. Owing to its computational efficiency, transparency, and long-standing application in geoscientific data analysis, K-means has been widely adopted for log-based facies and electrofacies classification in subsurface studies \cite{sadeghi2025clustering, di2014k, jain2010data}.
In this study, clustering was performed using the selected wireline log features described in Section~\ref{sec:desc}, with all variables standardised to zero mean and unit variance to ensure equal weighting in the distance-based optimisation.

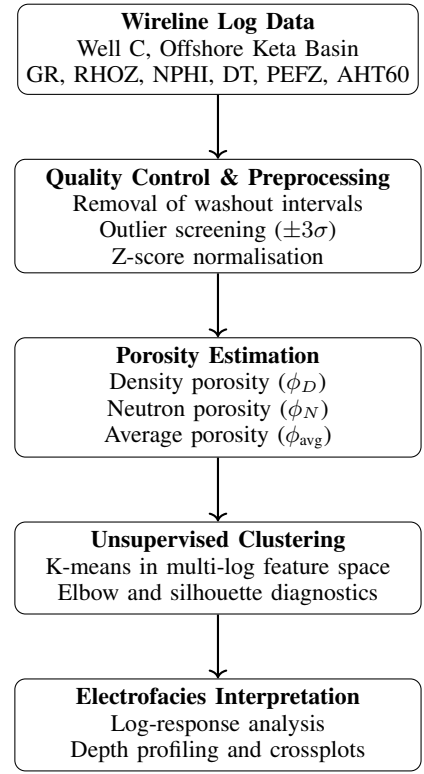
\begin{figure}[H]
\centering
\begin{tikzpicture}[
    node distance=0.85cm,
    every node/.style={font=\small},
    box/.style={rectangle, draw, rounded corners, align=center, minimum width=4.2cm, minimum height=1.1cm},
    bigbox/.style={rectangle, draw, rounded corners, align=center, minimum width=5.4cm, minimum height=1.2cm},
    arrow/.style={->, thick}
]

\node[box] (data) {%
\textbf{Wireline Log Data}\\
Well C, Offshore Keta Basin\\
GR, RHOZ, NPHI, DT, PEFZ, AHT60
};

\node[bigbox, below=of data] (qc) {%
\textbf{Quality Control \& Preprocessing}\\
Removal of washout intervals\\
Outlier screening ($\pm 3\sigma$)\\
Z-score normalisation
};

\node[bigbox, below=of qc] (poro) {%
\textbf{Porosity Estimation}\\
Density porosity ($\phi_D$)\\
Neutron porosity ($\phi_N$)\\
Average porosity ($\phi_{\text{avg}}$)
};

\node[bigbox, below=of poro] (cluster) {%
\textbf{Unsupervised Clustering}\\
K-means in multi-log feature space\\
Elbow and silhouette diagnostics
};

\node[bigbox, below=of cluster] (interpret) {%
\textbf{Electrofacies Interpretation}\\
Log-response analysis\\
Depth profiling and crossplots
};

\draw[arrow] (data) -- (qc);
\draw[arrow] (qc) -- (poro);
\draw[arrow] (poro) -- (cluster);
\draw[arrow] (cluster) -- (interpret);

\end{tikzpicture}
\caption{Schematic overview of the electrofacies analysis workflow applied to Well~C in the offshore Keta Basin. The pipeline integrates wireline log quality control, porosity estimation, and unsupervised K-means clustering to derive interpretable electrofacies in a data-scarce offshore setting.}
\label{fig:workflow}
\end{figure}

The number of clusters, $k$, was assessed using a combination of quantitative clustering diagnostics and geological considerations. The elbow method was applied to examine changes in within-cluster inertia as a function of increasing $k$, providing an initial indication of diminishing returns in variance reduction with additional clusters. Complementarily, silhouette analysis was employed to evaluate cluster cohesion and separation by measuring how well individual samples are assigned relative to neighbouring clusters \cite{rousseeuw1987silhouettes}. The distribution of silhouette coefficients across clusters was used to assess the internal consistency and interpretability of candidate clustering structures. Together, these diagnostics provide an objective basis for guiding cluster selection while allowing geological reasoning to inform the final electrofacies interpretation.

Figure~\ref{fig:workflow} summarises the complete analytical workflow adopted in this study. The pipeline proceeds from wireline log acquisition and quality control, through porosity computation and feature standardisation, to unsupervised clustering and electrofacies interpretation. The structure emphasises transparency and reproducibility, with each processing stage explicitly defined and independent of external training data.

\subsection{Electrofacies Label Assignment}

Electrofacies labels were assigned by interpreting the statistical clusters produced by unsupervised analysis within a geological framework grounded in established wireline log behaviour. The objective of this step was to translate multivariate clustering results into lithologically meaningful electrofacies based on rock matrix properties, rather than to infer fluid type or hydrocarbon saturation. This distinction is essential, as wireline logs primarily respond to mineral composition, clay content, porosity, and elastic properties unless explicit saturation modelling is performed \cite{tiab2024petrophysics, liu2023methods}.

Interpretation proceeded by examining the relative log-response characteristics of each cluster in the context of siliciclastic depositional systems. Gamma ray was used as the primary ordering parameter to establish a continuum of clay content, reflecting its sensitivity to potassium-bearing clay minerals and its widespread use in shale volume estimation \cite{vaidya2025principles, mondol2015well}. Clusters with higher mean GR responses were therefore interpreted as increasingly clay-rich, while lower GR responses indicated progressively cleaner, sand-dominated lithologies.

Additional discrimination between electrofacies was achieved by evaluating density, neutron, and sonic responses in combination. Elevated neutron porosity accompanied by lower bulk density and longer sonic transit times was interpreted as indicative of clay-bound water and reduced rock framework stiffness, characteristic of shale-dominated intervals. Conversely, lower neutron porosity, higher bulk density, and shorter sonic transit times reflect a more competent grain framework with reduced clay content, consistent with cleaner sandstone lithologies. The photoelectric factor provided supplementary mineralogical constraint, supporting differentiation between quartz-rich sands and clay-dominated units where log responses were transitional.

Based on these physically consistent log-response patterns, clusters were mapped onto a lithological continuum ranging from shale-dominated through mixed sand–shale electrofacies to cleaner sandstone-dominated electrofacies. These electrofacies represent statistically derived groupings interpreted in terms of rock type and texture, not predefined lithofacies or fluid classes. The resulting electrofacies distributions and their depth-dependent characteristics are presented and evaluated in the Results section.

\section{Results and Discussion} \label{sec:R4}

\begin{figure}
	\begin{minipage}{\linewidth}
		\centering
		\includegraphics[width=1.0\textwidth]{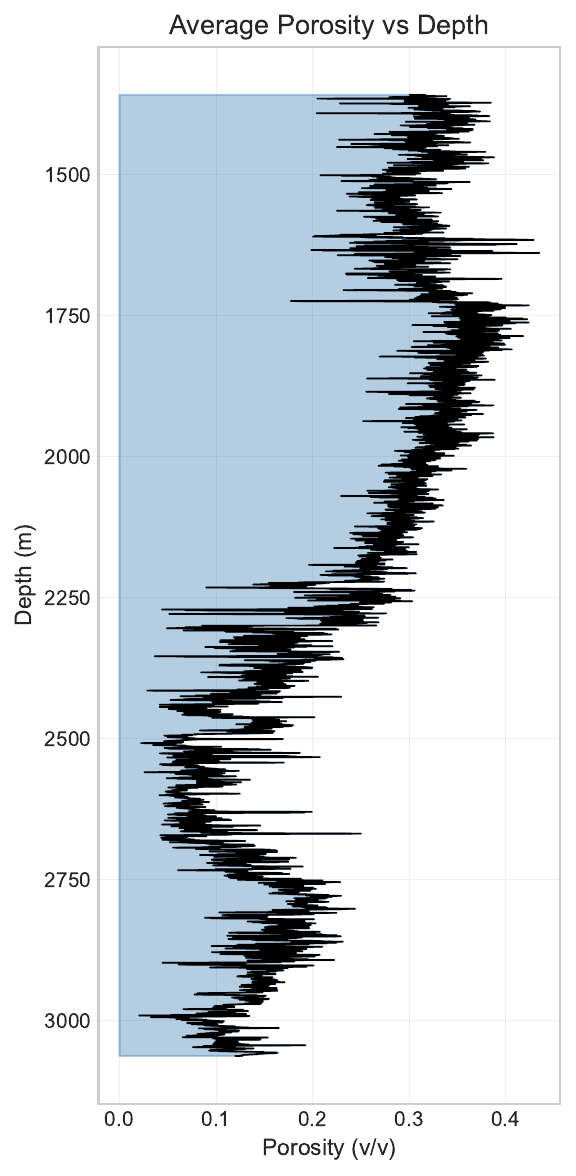} %
	\end{minipage}
	\caption{Average porosity ($\phi_{\text{avg}}$) derived from density and neutron logs plotted against depth for Well~C, illustrating compaction trends and intervals of relatively enhanced porosity.}
	\label{fig:fig2_porosity_depth}
\end{figure}

Figure~\ref{fig:fig2_porosity_depth} shows the vertical distribution of average porosity ($\phi_{\text{avg}}$) computed from density and neutron logs across the analysed interval of Well~C. Porosity exhibits an overall decreasing trend with depth, consistent with the progressive mechanical compaction and cementation that occur in siliciclastic successions. Superimposed on this background trend are several discrete intervals characterised by relatively elevated porosity, particularly within the mid-depth section of the well. These departures from the general compaction trend likely reflect variations in grain size, sorting, and clay content rather than fluid effects, given the absence of independent saturation calibration. The porosity profile therefore, provides a quantitative petrophysical baseline against which the subsequent clustering structure can be evaluated.

\begin{figure}
	\begin{minipage}{\linewidth}
		\centering
		\includegraphics[width=1.0\textwidth]{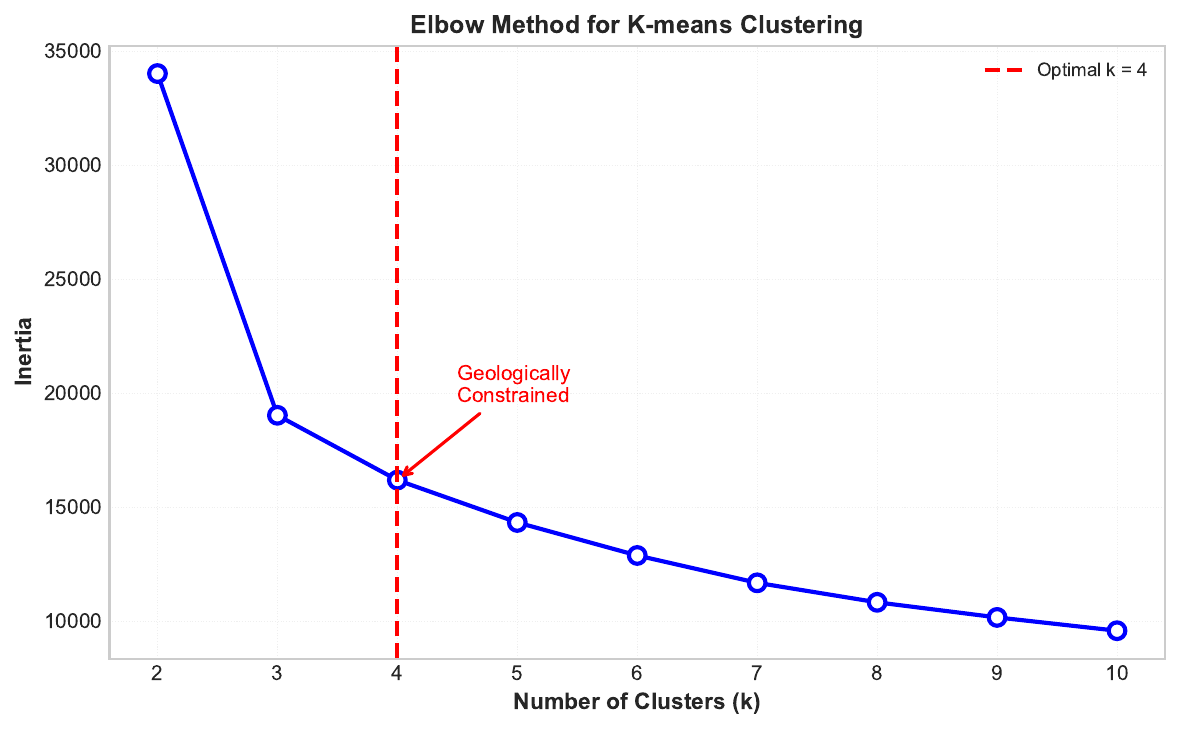} %
	\end{minipage}
	\caption{Elbow plot showing within-cluster inertia as a function of cluster number ($k$) for K-means clustering, highlighting diminishing variance reduction beyond moderate $k$ values.}
	\label{fig:Fig02_elbow_method}
\end{figure}
\begin{figure}
	\begin{minipage}{\linewidth}
		\centering
		\includegraphics[width=1.0\textwidth]{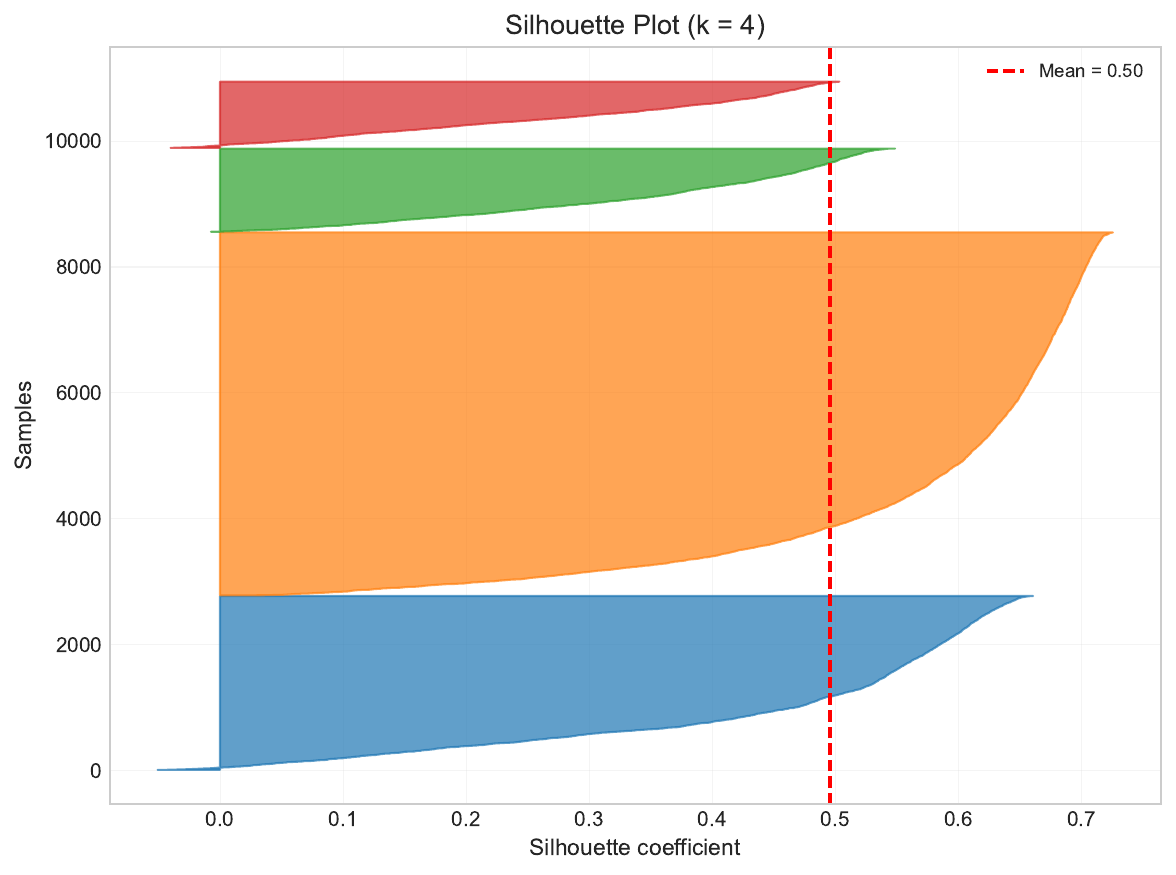} %
	\end{minipage}
	\caption{Silhouette plot for the selected clustering solution, showing the distribution of silhouette coefficients across clusters and the overall mean silhouette score.}
	\label{fig:fig4_silhouette}
\end{figure}

The robustness of the K-means clustering solution is assessed using complementary statistical diagnostics shown in Figs.~\ref{fig:Fig02_elbow_method} and~\ref{fig:fig4_silhouette}. The elbow plot (Fig.~\ref{fig:Fig02_elbow_method}) indicates a sharp reduction in within-cluster inertia between $k = 2$ and $k = 4$, followed by a more gradual decrease for higher values of $k$. This pattern suggests that the primary variance in the multivariate log space is captured by a small number of clusters, with additional clusters yielding diminishing improvements in compactness.

Silhouette analysis (Fig.~\ref{fig:fig4_silhouette}) provides a quantitative measure of cluster cohesion and separation. The average silhouette coefficient for the selected solution is approximately $0.50$, indicating moderate but meaningful separation between clusters. Most samples exhibit positive silhouette values, implying that they are, on average, closer to their assigned cluster centroid than to neighbouring clusters. The spread of silhouette values across clusters highlights the presence of transitional samples, which is expected in heterogeneous depositional environments characterised by gradual lithological changes rather than sharp boundaries. Taken together, the inertia and silhouette diagnostics indicate that the clustering solution captures statistically coherent structures in the data without excessive partitioning.

\begin{figure*}
	\begin{minipage}{\linewidth}
		\centering
		\includegraphics[width=1.0\textwidth]{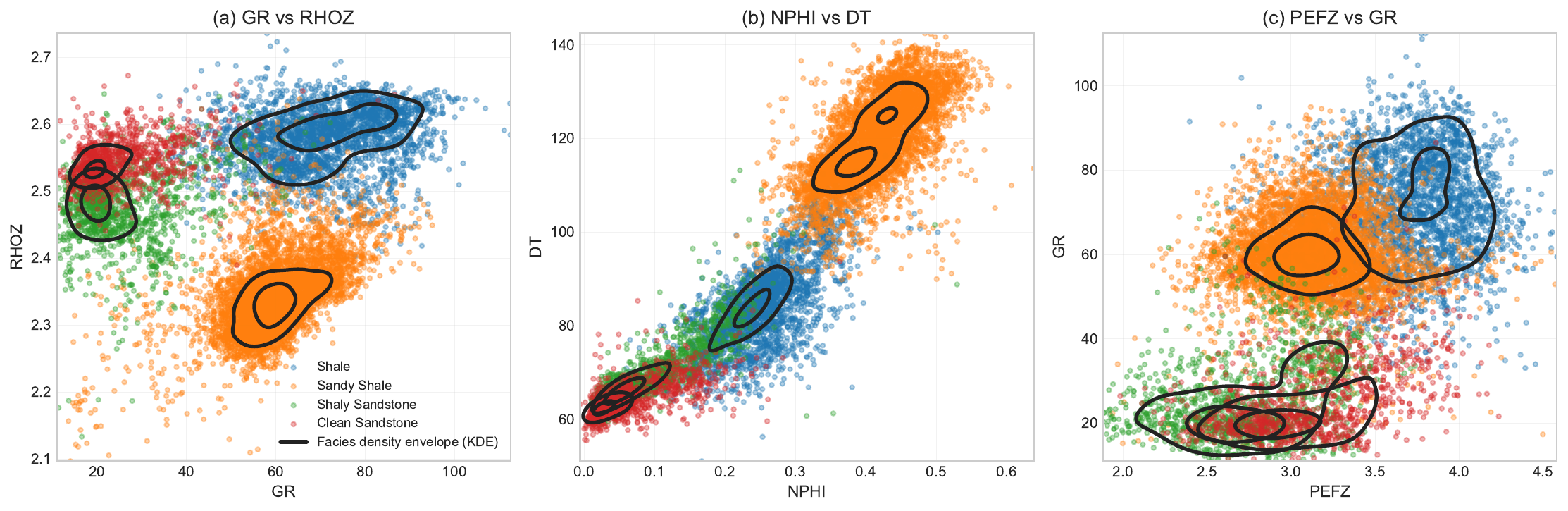} %
	\end{minipage}
	\caption{Composite log crossplots (GR--RHOZ, NPHI--DT, and PEFZ--GR) coloured by electrofacies, with kernel density envelopes indicating dominant cluster modes in multivariate log space.}
	\label{fig:fig5_composite_crossplots}
\end{figure*}

\begin{figure*}
	\begin{minipage}{\linewidth}
		\centering
		\includegraphics[width=1.0\textwidth]{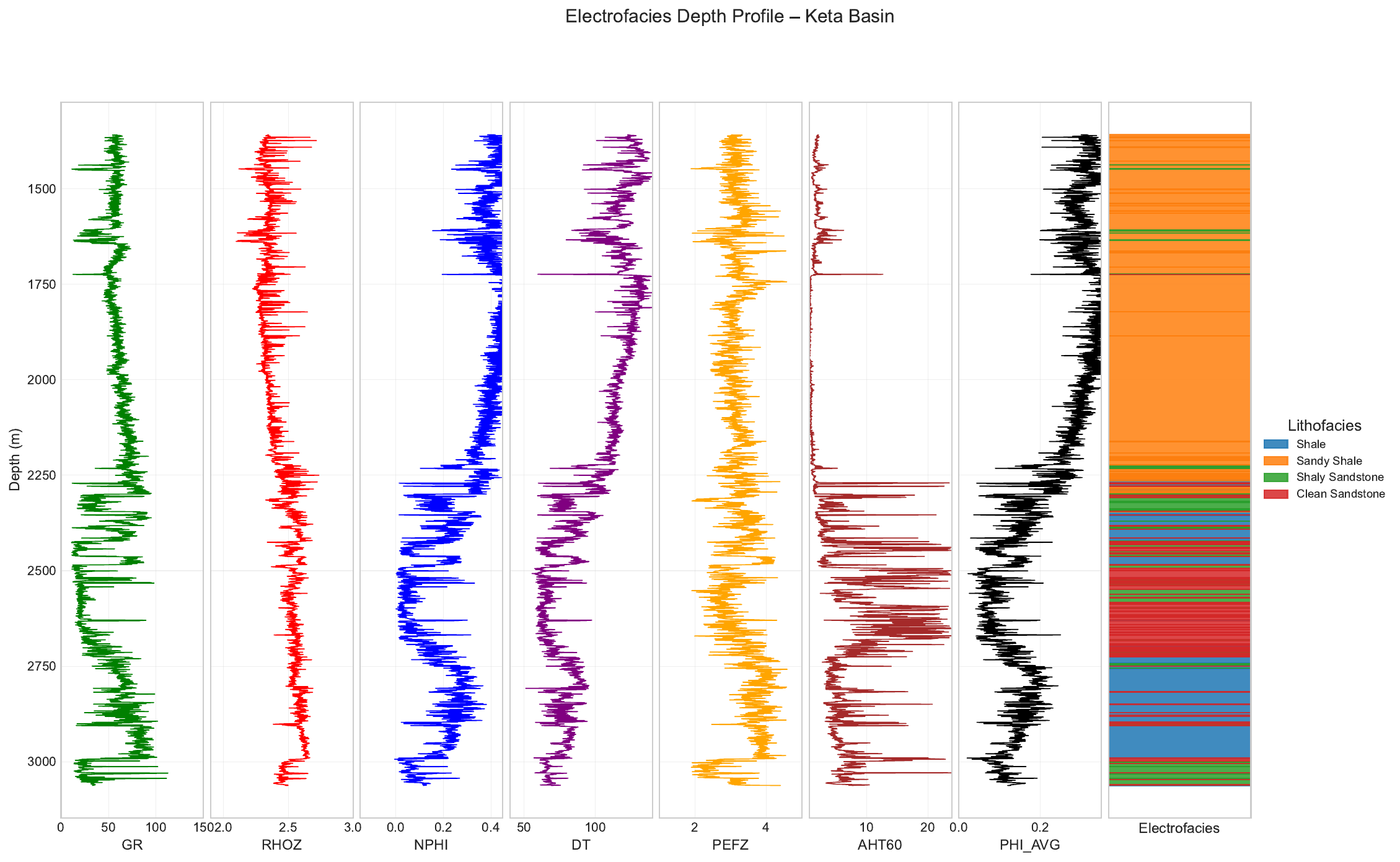} %
	\end{minipage}
	\caption{Electrofacies depth profile for Well~C, displayed alongside the wireline log suite and average porosity, illustrating the vertical continuity and petrophysical characteristics of each electrofacies.
}
	\label{fig:fig6_electrofacies_depth}
\end{figure*}

The statistical clustering structure is further examined through log crossplots shown in Fig.~\ref{fig:fig5_composite_crossplots}. In GR--RHOZ space, clusters occupy distinct regions corresponding to systematic variations in clay content and bulk density, with limited overlap relative to the full data spread. The NPHI--DT crossplot reveals a strong positive correlation consistent with porosity and elastic frame effects, along which clusters separate into partially overlapping but statistically distinct populations. In PEFZ--GR space, clustering highlights contrasts in mineralogical response, with photoelectric factor providing additional discrimination where gamma ray responses alone are ambiguous.
Kernel density envelopes superimposed on the scatter plots emphasise that clusters represent dominant modes in the joint probability distributions of the logs rather than arbitrary point groupings. Areas of overlap between density envelopes correspond to samples with intermediate log responses, which contribute to the moderate silhouette values observed. These crossplots therefore provide a statistical and physical explanation for the clustering behaviour identified by the quantitative diagnostics.

Figure~\ref{fig:fig6_electrofacies_depth} integrates the clustering results with the wireline log suite and average porosity profile to illustrate the vertical organisation of electrofacies in Well~C. Electrofacies occur as laterally continuous depth intervals rather than isolated points, indicating that the clustering is not driven by random noise or local artefacts. Intervals associated with higher average porosity generally coincide with electrofacies characterised by lower gamma ray responses and more compact sonic signatures, while lower-porosity intervals align with electrofacies exhibiting elevated gamma ray and neutron responses.
Transitions between electrofacies correspond to gradual shifts across multiple logs, consistent with the moderate silhouette separation observed in Fig.~\ref{fig:fig4_silhouette}. This behaviour supports the interpretation that the electrofacies reflect composite petrophysical states rather than sharply bounded lithological units. The depth profile thus provides a coherent synthesis of statistical clustering results and physical log responses, demonstrating the ability of the unsupervised workflow to organise complex wireline data into interpretable subsurface units.

\section{Conclusion}

This study developed and applied a reproducible, log-only workflow for electrofacies characterisation using standard wireline data from the offshore Keta Basin. By integrating basic petrophysical evaluation with unsupervised K-means clustering, the approach organises multivariate log responses into internally coherent and geologically interpretable electrofacies without reliance on core or laboratory measurements. The resulting electrofacies framework captures systematic variations in clay content, porosity, and rock framework properties observed along the analysed interval.

Four electrofacies were identified along a lithological continuum from shale-dominated to cleaner sandstone-dominated units. Their vertical distribution is consistent with expected variations in depositional energy and sediment composition within a heterogeneous siliciclastic system. Quantitative evaluation using silhouette analysis indicates moderate but meaningful cluster separation, supporting the robustness of the identified electrofacies while preserving transitional behaviour inherent in natural depositional environments.

The proposed workflow is particularly suited to frontier offshore basins where data availability is limited and rapid, cost-effective subsurface characterisation is required. Its reliance on widely available wireline logs and transparent statistical diagnostics makes it readily transferable to similar settings. Future work will focus on extending the methodology to multiple wells and integrating seismic attributes to evaluate lateral electrofacies continuity and support broader subsurface interpretation.

\section*{Acknowledgment}
The authors gratefully acknowledge the anonymous reviewers for their constructive and insightful comments, which contributed to improving the clarity and overall quality of this manuscript. Prof.~Atemkeng also acknowledges the financial support provided by Rhodes University, South Africa, towards the publication of this work.

\bibliographystyle{IEEEtran}
\bibliography{refs}

\end{document}